\documentclass[manuscript,screen]{acmart}%anonymous,review,
\AtBeginDocument{%
  \providecommand\BibTeX{{%
    \normalfont B\kern-0.5em{\scshape i\kern-0.25em b}\kern-0.8em\TeX}}}

\setcopyright{acmlicensed}
\copyrightyear{2024}
\acmYear{2024}
\acmConference[FAccT '24]{ACM Conference on Fairness, Accountability, and Transparency}{June 3--6, 2024}{Rio de Janeiro, Brazil}
\acmBooktitle{Proceedings of the ACM Conference on Fairness, Accountability, and Transparency (FAccT '24), June 3--6, 2024, Rio de Janeiro, Brazil}
\acmDOI{}

\usepackage{subcaption}

\begin{document}

% \title{Eliciting Personality in Large Language Models: A Text-based Big Five Personality Trait Classifiers Approach}

\title{Eliciting Personality Traits in Large Language Models}
% \title{Eliciting Big Five Personality Traits in Large Language Models: A Textual Analysis with Classifier-Driven Approach}
\renewcommand{\shorttitle}{Eliciting Personality Traits in Large Language Models}

\author{Airlie Hilliard}
\authornote{Both authors contributed equally to this research.}
\email{airlie.hilliard@holisticai.com}
\author{Cristian Munoz}
\authornotemark[1]
\email{cristian.munoz@holisticai.com}
\affiliation{%
  \institution{Holistic AI}
  \streetaddress{18 Soho Square}
  \city{London}
  \country{The United Kingdom}
  \postcode{W1D 3QH}
}

\author{Zekun Wu}
\email{zekun.wu@holisticai.com}
\affiliation{%
  \institution{Holistic AI}
  \streetaddress{18 Soho Square}
  \city{London}
  \country{The United Kingdom}
  \postcode{W1D 3QH}
}
\affiliation{%
  \institution{University College London}
  \streetaddress{Gower Street}
  \city{London}
  \country{The United Kingdom}
  \postcode{WC1E 6BT}
}
\author{Adriano Soares Koshiyama}
\affiliation{%
  \institution{Holistic AI}
  \streetaddress{18 Soho Square}
  \city{London}
  \country{The United Kingdom}
  \postcode{W1D 3QH}
}

\renewcommand{\shortauthors}{Airlie and Cristian, et al.}

\begin{abstract}
 Large Language Models (LLMs) are increasingly being utilized by both candidates and employers in the recruitment context. However, with this comes numerous ethical concerns, particularly related to the lack of transparency in these "black-box" models. Although previous studies have sought to increase the transparency of these models by investigating the personality traits of LLMs, many of the previous studies have provided them with personality assessments to complete. On the other hand, this study seeks to obtain a better understanding of such models by examining their output variations based on different input prompts. Specifically, we use a novel elicitation approach using prompts derived from common interview questions, as well as prompts designed to elicit particular Big Five personality traits to examine whether the models were susceptible to trait-activation like humans are, to measure their personality based on the language used in their outputs. To do so, we repeatedly prompted multiple LMs with different parameter sizes, including Llama-2, Falcon, Mistral, Bloom, GPT, OPT, and XLNet (base and fine tuned versions) and examined their personality using classifiers trained on the myPersonality dataset.  Our results reveal that, generally, all LLMs demonstrate high openness and low extraversion. However, whereas LMs with fewer parameters exhibit similar behaviour in personality traits, newer and LMs with more parameters exhibit a broader range of personality traits, with increased agreeableness, emotional stability, and openness. Furthermore, a greater number of parameters is positively associated with openness and conscientiousness. Moreover, fine-tuned models exhibit minor modulations in their personality traits, contingent on the dataset. Implications and directions for future research are discussed. \footnote{The Code, Models, and Dataset will be released upon acceptance.}
\end{abstract}




%, Dataset Influence on Model Output

%  \includegraphics[width=\textwidth]{figures/teaser.png}
%  \caption{System architecture for deriving personality profiles from large language model responses using text-based classification}


\maketitle

\section{Introduction}
The past decade has seen advancements in the way that personality is measured, with a number of innovative, technology-enabled approaches being proposed. Indeed, image-based assessments \cite{hilliard2022measuring}, smartphone data \cite{montjoye2013predicting}, eye movement tracking \cite{hoppe2018eye}, non-verbal behaviour in vlogs \cite{biel2011you}, and features extracted from Facebook profiles \cite{kosinski2013private} have recently been used to predict personality. These technology-enhanced solutions reduce the need for traditional self-report, which is associated with social desirability bias or faking, particularly in high-stakes contexts \cite{arthur2010magnitude}. Others have taken a more linguistic approach, measuring personality through the language used in YouTube videos \cite{biel2013hi}, social media posts \cite{park2015automatic}, blog posts \cite{YARKONI2010363} and video interviews \cite{hickman2021developing}. Of course, using language to measure personality is nothing new - in fact, the Big Five structure of personality emerged from linguistic analysis \cite{digman1990personality} and early models of personality were based on factor analysis of the language used to describe behaviour \cite{cattell1947confirmation}. However, more recent approaches to measuring personality combine language analysis with AI-driven techniques like natural language processing (NLP), a computational technique to analyse and interpret human language \cite{chowdhary2020natural}, to rapidly and automatically measure personality \cite{BOYD201763}. Such algorithmic language model-based approaches to measuring personality are widely being deployed in contexts like recruitment, where the personality of applicants can be measured through their answers to asynchronous video interviews \cite{hickman2019validity}\cite{hickman2021developing}, for example.

Computational techniques can also be applied to the generation of text, with artificial intelligence (AI) being used in applications such as to complete sentences, answer questions, and correct grammar \cite{brown2020language}. Indeed, the public release of models such as GPT with user-friendly interfaces (e.g., ChatGTP, Bard, Claude, etc.) marked an inflection point, with powerful models now at the fingertips of everyone rather than just those with programming skills. While the applications of these tools are vast, the power of chatbots is increasingly being harnessed in recruitment, where they are deployed to interact with applicants for tasks such as answering questions and screening applications \cite{nawaz2019artificial}. Here, NLP can be used to provide context and allow applicants to ask successive questions, which are then responded to by the bot, some of which can generate their own response \cite{bohmintent}. Their power is also being harnessed by applicants, who are using it for tasks such as resume personalisation \cite{kale2023job} and to prepare responses for interview questions e.g., \cite{Nolan_2023}. Others have even experimented with using chatbots to infer personality \cite{rao2023can} \cite{fan2023well}.

 With the complex nature of these algorithms, the use of conversational AI in recruitment has raised some concerns about how it can be applied in a responsible and ethical way, with some questioning the explainability and transparency of AI-driven recruitment tools since they are often black-box \cite{tippins2021scientific} \cite{hunkenschroer2022ethics}, meaning the internals of the model are uninterpretable or unknown \cite{guidotti2018survey}. Indeed, many of these AI-driven text-generation tools rely on artificial neural networks, systems that utilise parallel and connected processors to represent and process information in a structure that is said to resemble the structure of neurons in the human brain \cite{jain1996artificial}.
 Since neural networks are often complex and have a large number of connections \cite{setiono2000opening}, it is difficult to fully explain the model even if the input and outputs are known. This has implications for candidates since the black-box nature of systems may impact an employer's ability to communicate the capabilities and purpose of algorithmic systems \cite{kazim2021high}. 
 
 As a result, a body of research has emerged investigating how black-box systems can be made more transparent (see \cite{guidotti2018survey} for an overview), including the Prospector approach which aims to increase explainability by varying inputs and observing the effect on the output \cite{krause2016interacting}. Given the current increase in applications of LLMs in critical applications such as recruitment, the personality traits of LLMs could have implications for candidates using these tools to generate responses in preparation for interviews where the personality of the LLMs may be inferred by interviewees based on the language used in responses, rather than the personality of candidates themselves, particularly if the generated responses are used verbatim or with few edits. 

% As such, this study seeks to answer the following questions:
% \begin{itemize}
%     \item Can the personality profile of LLMs be elicited using prompts?
%     \item How do the personality profiles of LLMs vary by the number of parameters in the models?
%     \item Can LLMs be prompted to be higher in particular personality traits in accordance with the trait-activation theory?
% \end{itemize}

\begin{figure}[h]
\captionsetup{skip=2pt}
%\vspace{.3in}
\includegraphics[width=0.65\paperwidth]{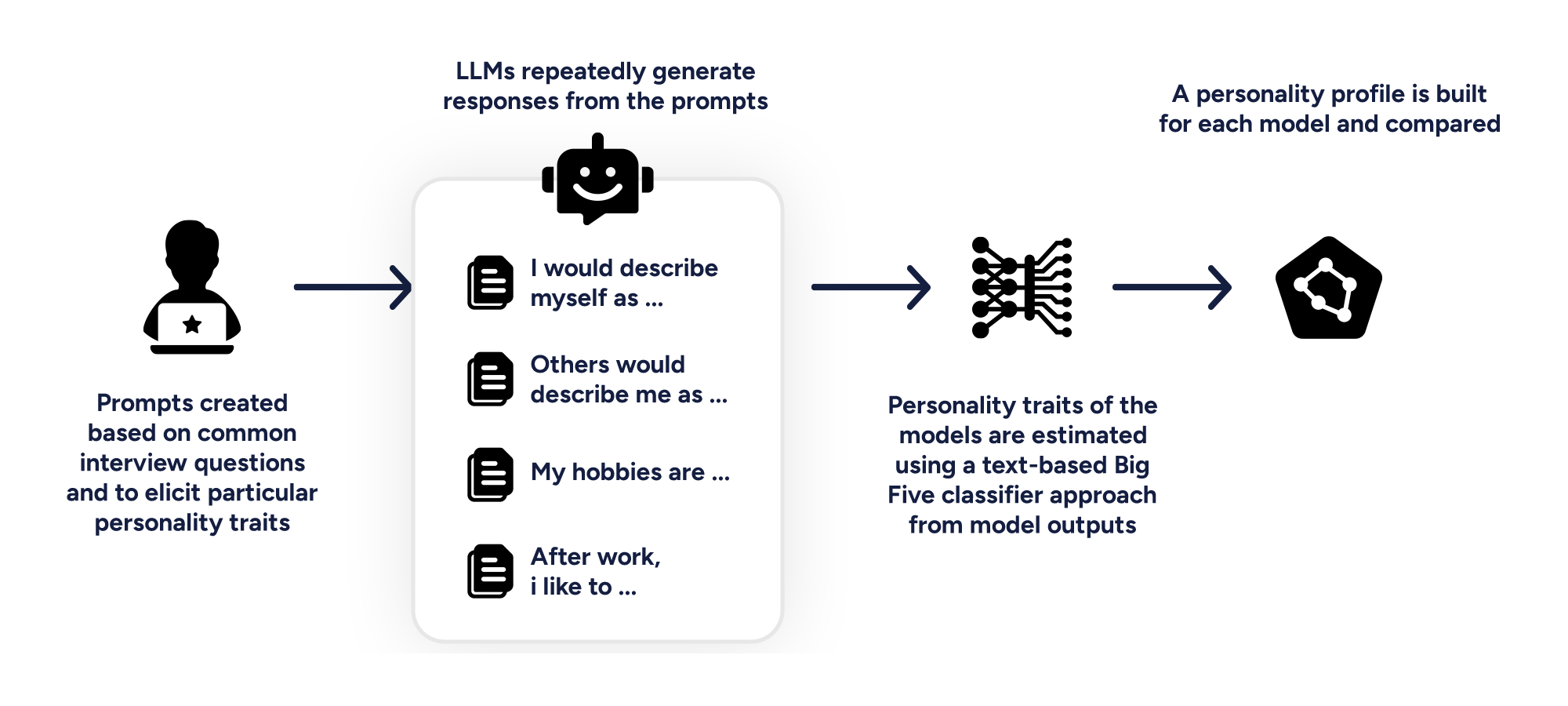}
%\vspace{.3in}
\captionsetup{skip=4pt}
\caption{System architecture for deriving personality profiles from large language model responses using text-based classification}\label{fig:text_gen_pipeline}
\end{figure}

In this study, we measure the Big Five personality traits of LLMS based on their responses to prompts derived from standard interview questions, as well as prompts designed to elicit high levels of specific traits. Although previous work (e.g.,  \cite{karra2022ai}; \cite{safdari2023personality})  has sought to measure the personality of LLMs by providing them with personality scales to respond to, this study is the first, to our knowledge, to infer the personality from LMs using the language from their outputs, analogous to how interviewers may intuitively use candidate responses to judge their personality in interviews (see Figure \ref{fig:text_gen_pipeline}). Specifically, we select LMs and provide them with prompt statements to complete: typical interview questions (tell me about yourself, strengths and weaknesses, etc.) and trait-activating questions designed to elicit higher levels of a particular Big Five trait. Based on these responses, we then use fine-tuned text classifiers to measure the personality of the LLMs. 

% Areas for future research include comparing the neural networks to a human sample. Limitations of the study are discussed.  

The remainder of this paper begins by giving an overview of the measurement of personality in job interviews and why it is an important construct to measure. We then describe linguistic analysis techniques and the neural networks that underpin LMs before describing the method and reporting our results. We find that:

\begin{itemize}
    \item All neural networks show high level of agreeableness and medium levels in other traits, except extraversion, which is slightly lower.
    \item Language Models with larger parameter size were trained with higher traits score for agreeableness, openness and emotion stability compared with smaller Language Models.
    \item Although larger models exhibit a broader range of personality traits, for conscientiousness, a execption of GPT, all the models are unaffected by trait-activating prompts. GPT3.5 and 4 presents sliglhy personality changes given the differnet inputs of standard questions. DIfferent to all the other models, GPT4 can handle Extraversion trait score  with minimal changes given any input, that is someting complete different to the others LM that have a great range variation.
    \item Agreeableness and Extraversion are the personality traits with major variation degree in all the models, execept of ChatGPT familiy, that increase their the variation of their personality trait for Agreeableness, Openess and conscientiousness.
    
    % likely because this is a social cue not taken into account by the models
\end{itemize}

% Implications and further directions for research are discussed.

%github figure is in the figures folder 

\section{Related Works}

This section provides an overview of existing literature on the Big Five personality traits in job interviews, text analysis and generation techniques, and the study of personality traits in LLMs.

\subsection{Personality and Job Interviews}
A large body of literature has established that the Big Five personality traits (conscientiousness, openness to experience, extraversion, agreeableness and neuroticism, which can be reversed to emotional stability) are among the strongest predictors of future job performance \cite{barrick1991big}\cite{schmidt1998validity}\cite{schmidt2016validity}\cite{schmitt2014personality}\cite{kuncel2010individual}\cite{higgins2007prefrontal}\cite{judge1999big}. As such, an applicant's personality is a significant factor to be taken into consideration when making hiring decisions. Although there are specific tools that can be used in recruitment to measure the personality of applicants - whether this be through self-reported, questionnaire-based methods such as the International Personality Item Pool (IPIP; \cite{goldberg1999broad}) or the Revised NEO Personality Inventory, or through image-based assessments \cite{hilliard2022measuring} - personality can also be inferred through structured job interviews \cite{levashina2014structured}. Indeed, common interview questions can be used to infer personality, particularly conscientiousness \cite{cortina2000incremental}, which is judged to include attributes like persistence, dependability and responsibility by hiring managers \cite{huffcutt2001identification}and is the most valid trait for predicting job performance across occupations \cite{barrick1991big}\cite{higgins2007prefrontal}\cite{judge1999big} \cite{kuncel2010individual} \cite{schmitt2014personality}. 

One explanation for the ability of hiring managers to infer personality from interviews  is offered by the trait-activation theory, which posits that personality traits are expressed at a greater rate when there are trait-relevant situational cues, or if there is an opportunity for trait expression \cite{tett2000situation}\cite{tett2003personality}. As such, given that many interview questions are designed to measure aspects of conscientiousness \cite{cortina2000incremental} by asking about achievements, future goals and motivation, this can explain why job interviews can be used to infer the levels of conscientiousness of candidates. According to the trait-activation theory, then, it is possible to elicit other traits during job interviews by altering the content of the interview to elicit trait-relevant cues. In other words, questions can be included in job interviews that elicit each of the Big Five traits, with prior research investigating the use of trait-activating questions  alongside automated personality analysis in language \cite{holtrop2022exploring}\cite{hickman2021automated}, as well as trait-activation in assessment centres \cite{lievens2006large} \cite{speer2015assessment}. 

\subsection{Text Analysis and Generation}

Text analysis is categorized into closed and open vocabulary approaches. Closed vocabulary methods like LIWC use predefined lists to predict personalities from online platforms \cite{boyd2022development}\cite{biel2013hi}\cite{golbeck2016predicting}, while the general inquirer focuses on concepts like power and wellbeing in Twitter personality predictions \cite{stone1962general}\cite{golbeck2011predicting}\cite{eichstaedt2021closed}. Conversely, open vocabulary approaches in NLP use algorithms to create word vectors, identifying word clusters from data for predictions \cite{eichstaedt2021closed}, with LSA and LDA being key examples, applied in personality assessment and language analysis \cite{landauer1997solution}\cite{kwantes2016assessing}\cite{garcia2014dark}\cite{blei2003latent}\cite{liu2016pt}\cite{schwartz2013toward}\cite{schwartz2013personality}.

%Text analysis can be categorized into closed and open vocabulary approaches. Closed vocabulary methods utilize predefined dictionaries or word lists representing specific constructs. For example, the Linguistic Inquiry and Word Count (LIWC) tool categorizes words into groups like personal pronouns, positive emotions, and verbs, and is utilized to predict personalities from platforms like YouTube and social media \cite{boyd2022development}\cite{biel2013hi}\cite{golbeck2016predicting}. Another instance is the general inquirer, updated with concepts related to power and wellbeing, predicting personality from Twitter data \cite{stone1962general}\cite{golbeck2011predicting}\cite{eichstaedt2021closed}.

%Open vocabulary approaches, typical in natural language processing, create relational word vectors. These algorithms identify word clusters from data, facilitating predictions \cite{eichstaedt2021closed}. Latent semantic analysis (LSA) and Latent Dirichlet Allocation (LDA) are primary examples of this approach, with applications in measuring personality from essays, social media posts, and analyzing language use variations \cite{landauer1997solution}\cite{kwantes2016assessing}\cite{garcia2014dark}\cite{blei2003latent}\cite{liu2016pt}\cite{schwartz2013toward}\cite{schwartz2013personality}.

Contrarily, Natural Language Generation (NLG) synthesizes comprehensible text from data. It involves steps like content determination, sentence planning, and realization for grammatically correct outputs \cite{reiter1997building}. Contemporary NLG techniques harness deep learning, employing neural networks like the Encoder-Decoder framework for text production \cite{gatt2018survey}. A notable example is BERT, which leverages bi-directional text representations, differing from prior tools focusing only on left-sided data. Its methodology includes pre-training on unlabelled content and subsequent fine-tuning using labeled data \cite{devlin2018bert}.

\subsection{Personality Traits of Large Language Models}
Given that LLMs are being used to generate responses to anticipated job interview questions, a trend that is increasingly being driven by social media \cite{Nolan_2023}, this could have implications for how candidates are evaluated in job interviews. This is particularly the case if applicants do little to no editing or customisation of the responses provided by the LLMs, wherein inferences made about a candidate's conscientiousness, for example, could be affected by the personality of the LLMs used to generate the planned answer. This could, therefore, influence the way that applicants are perceived by potential employees and have implications for their hireability. Therefore, this study aims to investigate the personality traits of LLMs from their responses to interview questions, using both common interview questions and those specifically designed to activate each of the personality traits. Specifically, we build on previous attempts to measure the personality of LLMs but contextualise this to recruitment in the current study. 

For example,  Karra and colleagues \cite{karra2022ai} investigated the personality of multiple LLMs (GPT-2, GPT-3, Transformer XL, and XLNet). They provided the models with prompts in the form of statements from a Big Five personality questionnaire, and used the language models to generate responses. Using a zero-shot classifier \cite{yin2019benchmarking} \cite{xian2018zero} to analyse the text, the probability score for each Big Five trait was transformed to measure the personality of the language model on a scale of 1-5. They found that GPT-3 is the highest in agreeableness and TransformerXL is the highest in conscientiousness, with around median levels of the other traits across all models. Similarly, Serapio-Garcia et al., \cite{safdari2023personality} provided LLMs from the PaLM family with prompts to rate items in the IPIP-NEO and Big Five Inventory based on persona descriptions to establish validity. They then investigated whether the personality scores of LLMs could be shaped using linguistic qualifiers and trait adjectives, finding that both attempts at single-trait shaping and mixed-trait shaping were effective at changing the personality scores of the models. This study also highlighted the superior reliability and validity of synthetic LLM personality in larger, instruction fine-tuned models compared to smaller, non-instruction-tuned ones. Our research approach echoes this by incorporating models of both types and varying sizes. Unlike this study that focused solely on Flan-PaLM(540B, 62B, 8B instruction fine-tuned) and PALM(62B, non-instruction-tuned), we extended our analysis to a wider range of LLMs. 

However, another recent piece of research indicates that LLM responses to personality tests cannot be interpreted in the same way as human responses, where LLM responses systematically deviate from typical human responses. For example, LLMs respond to positively and negatively framed statements in the same way, whereas humans would be expected to respond negatively to the reverse-coded item \cite{solar2023}. Furthermore, when LLMs are promoted towards particular personality traits, there is a lack of a clear five-factor structure that is seen in equivalent human attempts. As such, the present study aims to investigate the personality of LLMs through the language they use, rather than through human-aimed personality inventories and build on existing research to investigate whether LLMs respond to trait-activation in the context of interviews.

% \begin{itemize}
%     \item \textbf{Use of Sentence Prompts}: Instead of providing models with complete sentences as seen in other studies, we give them sentence prompts. This method allows models to respond more freely, revealing their natural tendencies without being influenced by a given context.

%     \item \textbf{Emphasis on Interview Questions}: Most existing research relies on standard personality questionnaires. In contrast, we center our study on prompts derived from interview questions. This approach provides insights into how models might respond in job interview situations, offering a view of their apparent personality in practical scenarios.

%     \item \textbf{Broad Range of Models}: We don't limit ourselves to a few specific models. While previous studies focus mainly on a few models, we include a wider variety of both fixed instruction fine-tuned and non-instruction-tuned LLMs for a more in-depth analysis.

%     \item \textbf{Analysis Using Trait-activating Prompts}: Our study assesses how models respond to both standard and trait-activating prompts. This helps us understand if a model's personality changes or remains stable when presented with prompts targeting specific traits.

%     \item \textbf{Regression-based Approach}: Instead of the usual practice of using a classifier's probability score to measure personality, we adopt a regression-based method. This technique offers a clearer and more accurate representation of a model's personality characteristics.

% \end{itemize}

\section{Methodology}

Building upon the foundation laid by prior works, this study adopts novel methodologies in the following ways:

\begin{itemize}
\item We incorporate a broader range of state-of-the-art LLMs, encompassing both specialized instruction/chat fine-tuned versions and the foundational base models, to ensure a comprehensive analysis.
\item Our elicitation prompts are carefully tailored to stimulate real-world job interview scenarios, directly tying our research to the context of recruitment.
\item Instead of using a traditional personality Question \& Answering inventory, we challenge the models with sentence completion tasks, which more accurately reflect natural language usage.
\item We employ the classification-based evaluation method that quantifies the models' personality traits by converting classifier probability scores into a continuous spectrum.
\item The models' responses to both standard and trait-specific prompts are compared, allowing us to assess the adaptability and depth of their personality representation.
\end{itemize}
% \begin{itemize}
%     \item We include extensive SOTA LLMs with both diverse version and type of both instruction/chat fine-tuned and base LLMs in our analysis.
%     \item We design prompts based on interview questions, contextualising the current study to recruitment scenario; 
%     \item We provide the models with sentence prompts to complete, rather than a personality inventory with Question Answering questions;
%     \item We adopt classification-based evaluation approach, transforming the probability score of classifiers to measure personality on a continuous scale;
%     \item We compare the personality of models when they complete standard and trait-activating prompts;
    
% \end{itemize}

\subsection{Large Language Models in Experiments}
This study utilizes autoregressive transformer models from the Commercial APIs and the Hugging Face library \cite{hugging_face_transformers_documentation}. We specifically chose autoregressive transformer models such as GPT, OPT, XLNet, Llama 2, and Falcon, and so on. The BERT series was not included because it functions as an autoencoder model, which differs from our focus \cite{illustrated_gpt2}. For token prediction, we employed a sampling decoding strategy, which inherently produces non-deterministic outputs. To improve the quality of these outputs, we implement the  hyper-parameter tuning to models.

\paragraph{\textbf{GPT:}} The GPT series by OpenAI is a progression of decoder-only language models. The series started with GPT-1, which had 117 million parameters and was trained on BooksCorpus \cite{Radford2018ImprovingLU}. It then expanded to GPT-2 with 1.5 billion parameters, trained on WebText \cite{radford2019language}. GPT-3 followed, with 175 billion parameters and training on datasets like Common Crawl \cite{brown2020language}. GPT-3.5 Turbo was introduced to enhance real-time performance, and the latest, GPT-4, boasts 1.76 trillion parameters and capabilities for multimodal tasks \cite{openai2023gpt4}.
\paragraph{\textbf{Llama 2:}} Developed by Meta AI, Llama 2 consists of autoregressive language models of various sizes (7B, 13B, 70B) \cite{touvron2023llama}. It is pretrained on a corpus of 2 trillion tokens and fine-tuned with human-annotated examples. Llama 2 is designed for both commercial and research applications and runs on Meta's Research Super Cluster and third-party cloud resources. Meta has offset its carbon footprint of 539 tCO2eq.
\paragraph{\textbf{Falcon:}} Falcon, created by the Technology Innovation Institute, includes a set of causal decoder-only models with sizes ranging from 7B to 180B \cite{falcon}. These models are pretrained on the RefinedWeb dataset \cite{penedo2023refinedweb} and demonstrate superior performance due to extensive training and optimized architectures featuring FlashAttention. Despite its size, Falcon-180B is designed for efficient inference and is commercially available under permissive licenses.
\paragraph{\textbf{Mixtral:}} The Mixtral series, developed by Mistral AI \cite{jiang2024mixtral}, features a range of decoder-only Sparse Mixture-of-Experts models, including the prominent Mixtral 8x7B and 7B, available in both base and instructor versions. These models combine a large total parameter count, reaching up to 46.7 billion, with efficient processing, utilizing only 12.9 billion parameters per token. With training on diverse datasets from the open web, these models surpass competitors like Llama 2 70B and GPT-3.5, particularly in inference speed.
\paragraph{\textbf{XLNet:}} XLNet is an autoregressive language model that addresses some of BERT's limitations in joint probability modeling \cite{yang2019xlnet}. It utilizes the Transformer-XL architecture, combining AR and AE features for improved performance on longer texts. XLNet has been trained on various datasets, including BooksCorpus \cite{zhu2015aligning}, Wikipedia, Giga5 \cite{parker2011english}, ClueWeb \cite{callan2009clueweb09}, and Common Crawl \cite{commoncrawl}. Its larger variant, 'XLNet large', has additional layers and size for better performance.
\paragraph{\textbf{OPT:}} The OPT suite is a collection of decoder-only transformer models, similar in size and performance to GPT-3 \cite{zhang2022opt}. It employs a casual language modeling (CLM) objective and has been pre-trained on datasets including BookCorpus, CC-Stories, The Pile, Pushshift.io Reddit dataset \cite{baumgartner2020pushshift,roller2021hash}, and CCNewsV2, which was also used in training RoBERTa \cite{liu2019roberta}. The OPT models come in three sizes: OPT-125m, OPT-350m, and OPT-1.3b.
\paragraph{\textbf{Additional models:}} In addition to the primary models, our analysis incorporates fine-tuned versions of GPT-2 and GPT-J-6B for specialized text generation tasks. These models simulate the language of celebrities or address controversial topics. The GPT-2 variants were fine-tuned on the styles of Shakespeare, Rihanna, Michael Jackson, Yann Lecun, and Elon Musk, while GPT-J-6B was tailored for Shakespeare, 4chan, and Shinen styles. The configurations were aligned with the primary models for consistency.

\subsection{Interview Prompt Design for Trait Elicitation}
Since the generators are designed to complete sentences rather than to answer questions, interview questions were reframed as prompt statements to be completed by the model. To investigate whether trait-activating questions had an effect on the personality of the model according to the language analysis, we provided the LLMs with both standard interview questions and trait-activating questions, with 5 prompts being created per trait/theme. For the standard interview questions, we created prompts in relation to tell me about yourself, cultural fit/ideal workplace, strengths and weaknesses, future plans (where do you see yourself in X years?), and coping under pressure since these are commonly asked interview questions \cite{smith_2022}\cite{oliver_2021}. For the trait-activating questions, questions were adapted from \cite{holtrop2022exploring} for some of the conscientiousness and extraversion prompts while for the remaining traits, prompts were created by adapting statements from the International Personality Item Pool (IPIP) scales \cite{goldberg1999broad}. There were, therefore, 25 questions per category (standard or trait-activating), or 50 in total. Prompts can be seen in Appendix \ref{ap:questions}. For each prompt, 1000 answers or completions were completed, resulting in 5000 text strings for each trait/theme since there were 5 questions. This process generated a cumulative total of 50,000 texts for each model. The maximum sentence length for text strings was set to 128 words.

\subsection{Classifier-Based Personality Analysis}
% We use personality analysis tools to evaluate text generated by LLMs. Our selected baseline methodology is the tool developed by Li \cite{personality_li}, intended for assessing Facebook users' personalities based on their status updates, is particularly suitable for our needs. This is due to its ability to analyze texts similar in length to neural network outputs. Although other models like the one by Mehta et al. \cite{mehta2020bottom}, trained on the Essays dataset for LIWC \cite{pennebaker1999linguistic}, are available, Li's tool aligns better with our data format. It was developed from the myPersonality project \cite{mypersonality}, which gathered personality assessments and Facebook status updates from users. This tool utilizes a random forest regressor and classifier for personality prediction, exploiting the text analysis strengths of random forest models (as discussed in \cite{sun2020application} \cite{xue2015research}). It outputs both continuous scores and binary outputs for personality traits. In the present study, however, the random forest regressor model, while replacing Facebook statuses with neural network outputs to analyze personality traits, demonstrated limitations in robustness and efficacy. As shown in Table \ref{tab:classifier_results}, the model's performance was not satisfactory for our experiments.

In this study, we evaluate text generated by LLMs using personality analysis tools. The baseline tool for our analysis, developed by Li \cite{personality_li}, originally assesses Facebook users' personalities from their status updates. Its capability to analyze text lengths comparable to LLM outputs makes it well-suited for our needs. Although Mehta et al.'s model \cite{mehta2020bottom} is an alternative, it is less compatible with our data format, being trained on a different dataset for Linguistic Inquiry and Word Count (LIWC) \cite{pennebaker1999linguistic}. Li's tool, derived from the myPersonality project \cite{mypersonality}, utilizes a random forest regressor and classifier for personality prediction. This method capitalizes on the text analysis capabilities of random forest models \cite{sun2020application} \cite{xue2015research}, providing both continuous scores and binary outputs for personality traits. However, when we adapted the random forest regressor model to replace Facebook statuses with LLM outputs, it showed limited robustness and efficacy, as demonstrated in Table \ref{tab:classifier_results}.

Due to the limitations of previous methods, we adopted a more advanced methodology, employing five transformer-based models. These include two BERT and three DistilBERT models, all fine-tuned using the MyPersonality dataset \cite{stillwell2015mypersonality}. The dataset provides binary labels for the Big Five personality traits - namely, extraversion, neuroticism, agreeableness, conscientiousness, and openness (cEXT, cNEU, cAGR, cCON, cOPN). We utilized it to refine five binary classifiers for text classification, each yielding a probability score reflecting the likelihood of a specific personality trait being present. To better align with recruitment preferences that often favor emotional stability over neuroticism, we modified the scoring approach. We calculated the Emotional Stability Score as 1 minus the neuroticism score, thus inversely representing emotional stability and offering a more nuanced analysis of personality traits.

% Given the inadequacy of the state-of-the-art (SOTA) random forest method in terms of robustness and efficacy for our experiments, we have transitioned to a more advanced approach. We now implement five transformer-based models, comprising two Bert and three Distilbert models, fine-tuned with the MyPersonality dataset \cite{stillwell2015mypersonality}. This dataset provides binary labels in the categories of cEXT, cNEU, cAGR, cCON, and cOPN, corresponding to the Big Five personality traits, which are used to fine-tune five binary classifiers under text classification downstream task. The output of each classifier is the probability score that represents the likelihood of being high or low in a trait. Moreover, given that emotional stability is often more desirable than neuroticism during recruitment, neuroticism scores were reversed to reflect emotional stability. This shift to specialized language models for binary classification aims to achieve a more detailed and nuanced understanding of personality traits in language, addressing the shortcomings of previous methodologies.

\begin{table}[ht] 
\caption{Performance F1 Score of classifier model}
\label{tab:classifier_results}
\begin{center}
\begin{tabular}{lccccc}
\toprule
& Openness & Conscientiousness & Extraversion & Agreeableness & Neuroticism \\
\midrule
BERT Based (Our) & 85.960 \% & 63.076 \% & 63.530 \% & 69.905 \% & 55.566 \% \\
Random Forest (Baseline) & 83.634 \% & 43.355 \% & 38.848 \% & 61.442 \% & 36.428 \% \\
\bottomrule
\end{tabular}
\end{center}
\end{table}

% To validate the classifier and make sure the classifiers make prediction align with human conmonssense reasoning, we use SHAP to explain the model classification. An SHAP example of how the classifiers work can be seen in Figure \ref{fig:shap-1} using an example output from llama 2-7b with a prompt designed to elicit high levels of openness to experience. The output was given a probability of 0.70 for high levels of conscientiousness. Here, the words "enjoy" and "challenging" in particular have a strong weighting from the classifier, which is logical considering that high levels of conscientiousness are characterised by self-efficacy and a focus on achieving goals. In other words, enjoying a challenge. For the same text, as shown in Figure \ref{fig:shap-2}, the agreeableness classifier results in a probability of .90, with appeasing picky eating and the unmatched bonding over food contributing to agreeableness, or the tendency to be cooperative and altruistic.

\begin{figure}[h]
\centering
\includegraphics[width=0.95\linewidth]{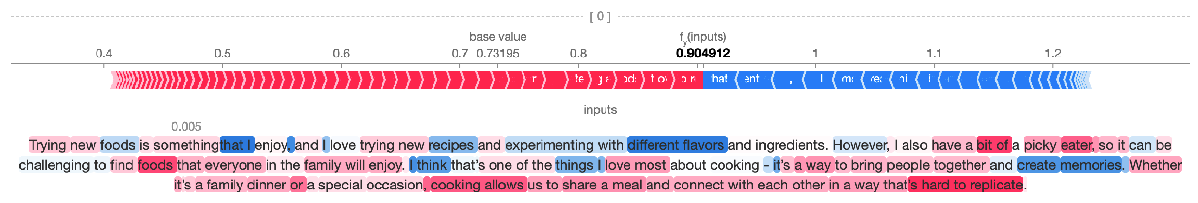}
\caption{SHAP visualization illustrating the classifier's rationale for agreeableness, with red indicating positive contribution and blue indicating negative contribution to the "yes" classification.}
\label{fig:shap-1}
\end{figure}

\begin{figure}[h]
\centering
\includegraphics[width=0.95\linewidth]{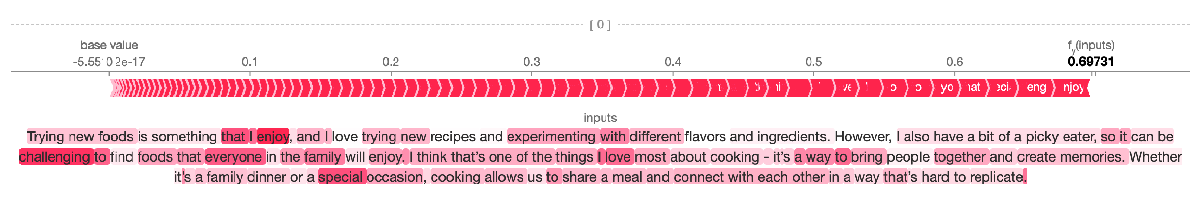}
\caption{SHAP visualization illustrating the classifier's rationale for conscientiousness, with red indicating positive contribution and blue indicating negative contribution to the "yes" classification.}
\label{fig:shap-2}
\end{figure}

To ensure our classifier's predictions are not only accurate but also intuitively correct, we employ SHAP (SHapley Additive exPlanations) to explain the model's decision-making process. Figures \ref{fig:shap-1} and \ref{fig:shap-2} provide SHAP visual explainability for our classifiers, where the color-coded contributions—red for affirmative influence and blue for negative—guide us in understanding the lexical elements that sway the classifiers toward a "yes" or "no" decision. The first SHAP analysis is depicted in Figure \ref{fig:shap-1}, where we dissect the model's reasoning based on an output from Llama2-7b, prompted to draw out a high degree of openness to experience. The model ascribed a 0.70 probability indicating strong conscientiousness. Notably, the terms "enjoy" and "challenging" were heavily weighted in the classifier's decision, aligning with the characteristics of conscientious individuals who are often driven by self-discipline and goal achievement—traits synonymous with relishing challenges.

Simultaneously, Figure \ref{fig:shap-2} reveals the classifier's logic for agreeableness, assigning a high probability of 0.90. Phrases pertaining to accommodating dietary preferences and the unique camaraderie formed through shared meals were interpreted as markers of agreeableness, reflecting the propensity for cooperation and kindness.

\section{Results}

For clarity in our comparative analysis, this study delineates language models into two categories: \textit{small language models}, including those with up to 1.3 billion parameters, and \textit{large language models}, comprising those with a parameter count exceeding the 1.3 billion threshold. This section will provide detailed insights into the text generation process and the ensuing analysis performed on personality trait scores, responding to both trait-activating and standard questions, across standard and fine-tuned models.

\textbf{Text Generation}: In this study, each question prompt yielded 1,000 responses, resulting in a total of 5,000 texts for each characteristic or topic, given that there were five unique questions. This process generated a cumulative total of 25,000 texts for each model. The tool used to generate these responses was obtained from Huggingface\footnote{https://huggingface.co}. The "Sampling" method was employed for generation, configured with a temperature setting of 1.0, a top-k value of 40, a top-p value of 0.95, and a maximum length of 128 tokens. During post-processing of the utterances, non-ASCII characters, repeated non-gram words greater than three at the end of text strings (when the model starts to generate the same words again and again) and words of over 20 characters that did not make sense were removed from the generated text before analysis. To evaluate the models we use the classifier probability output. To avoid offsets that can be generated related to the initial prompt we normalized the output using the following equation:

\begin{equation}
    \mathcal{N}(\textrm{score}_{trait}) = \textrm{score}_{trait}(\mathrm{sentence}) - \textrm{score}_{trait}(\mathrm{prompt}) + 0.5
\end{equation}

Therefore, we can see that when the model increases or decreases the characteristic score in relation to the initial prompt, we will have values below or above 0.5, respectively.

\begin{figure}
    \centering
    \includegraphics[width=0.95\linewidth]{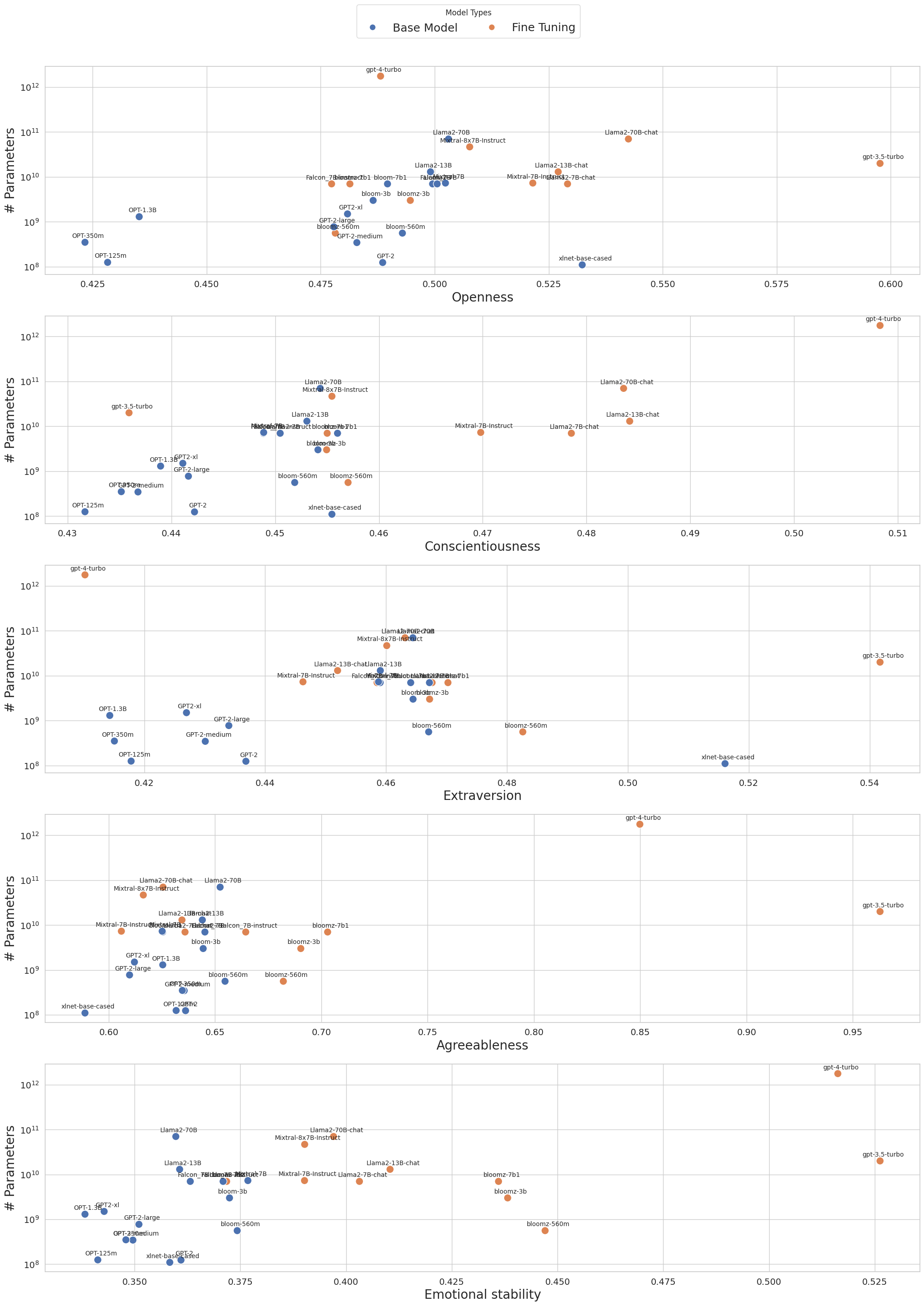}
    \caption{Comparison of Model Size and Personality Trait Scores 'Trait Activating Question' Datasets. Orange points represent the base model version, while the blue points represent the chat or instruct version.}
    \label{fig:model_size_vs_trait_score}
\end{figure}

\begin{figure}
    \centering
    \includegraphics[width=\linewidth]{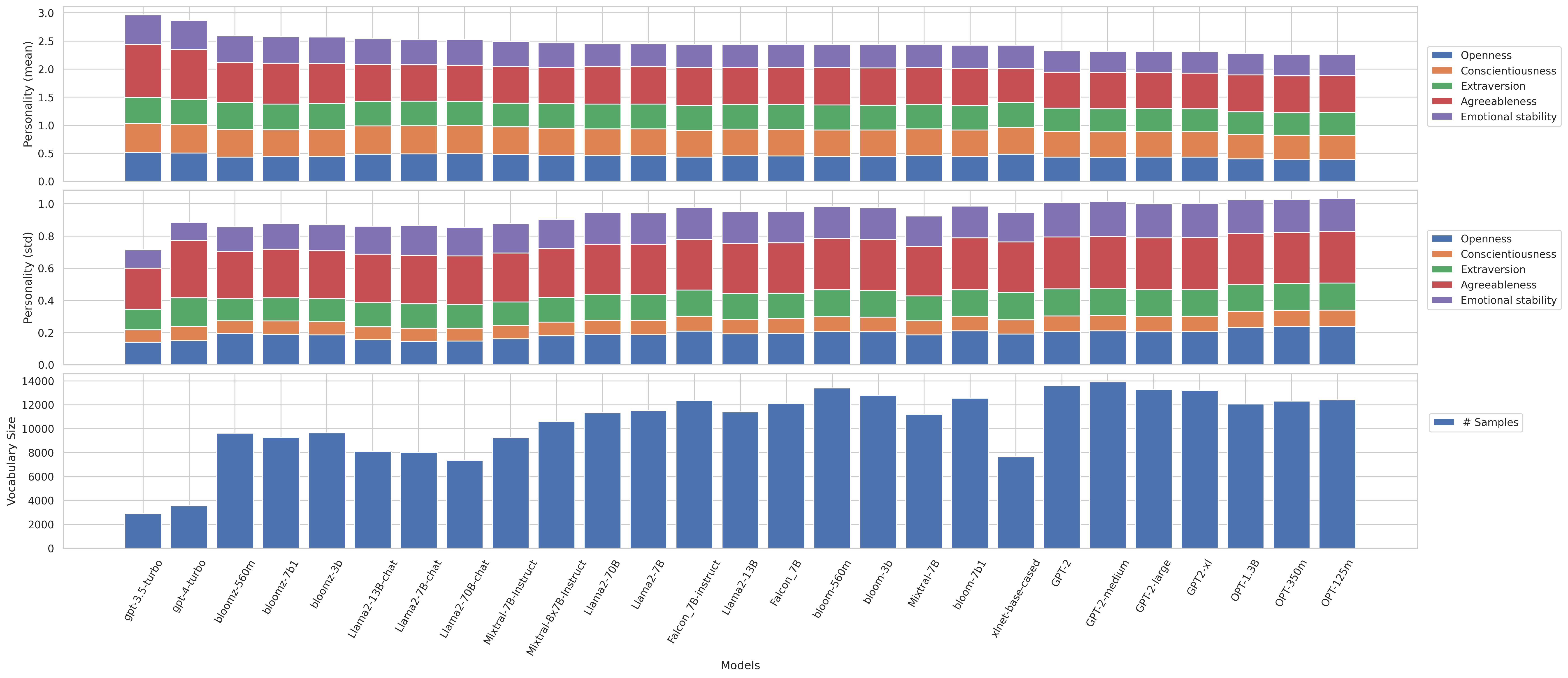}
    \caption{Average, Standard Deviation of Presonality Trait Score  and Vocabulary Size for each LLM.}
    \label{fig:combined}
\end{figure}

\begin{figure}
    \centering
    \includegraphics[width=0.8\linewidth]{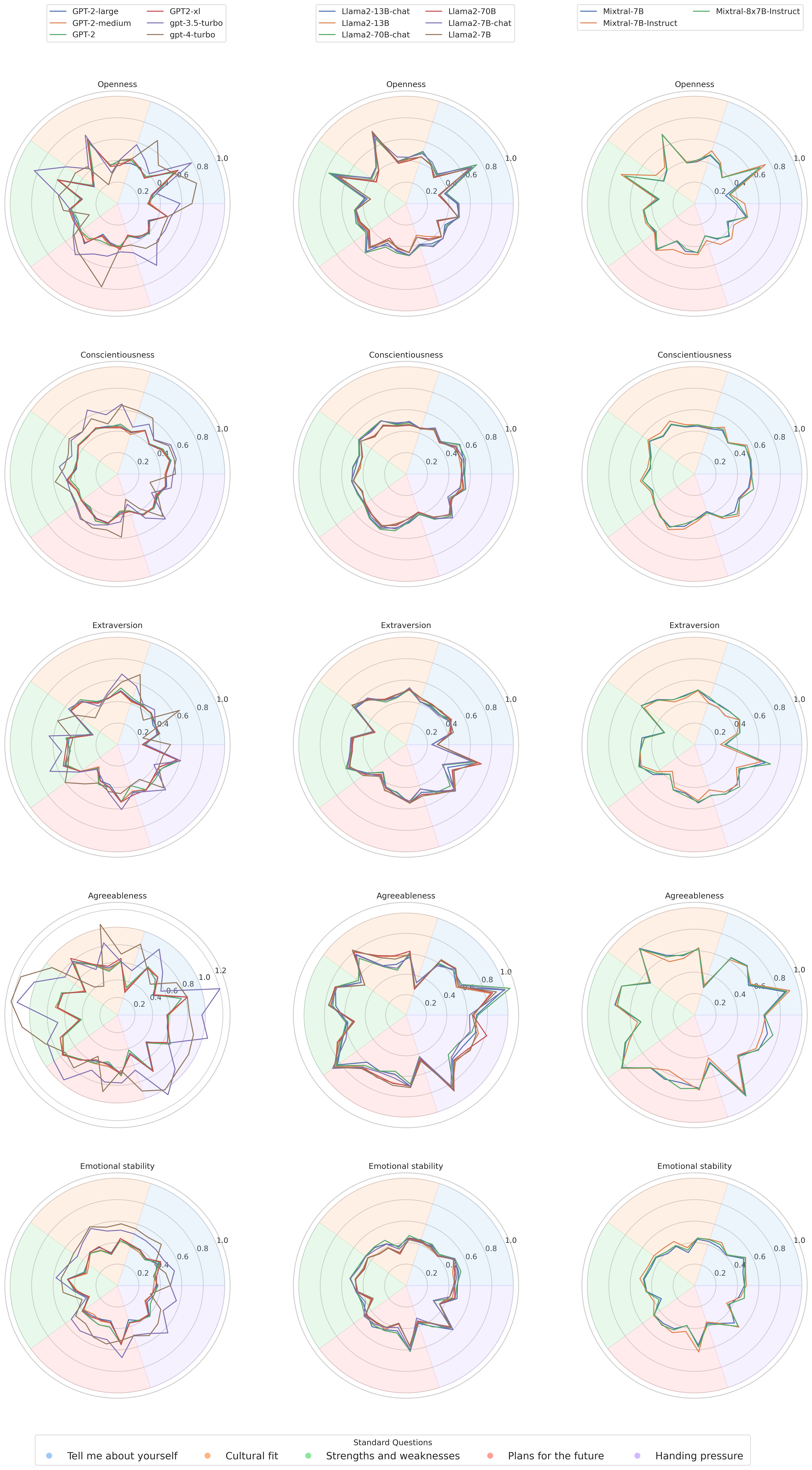}
    \caption{Comparison of the Mean Big Five Scores for each for categories of standard interview questions.}
    \label{fig:standard_questions_families}
\end{figure}

\textbf{Trait-Activating Questions}: The responses to the trait-activating questions were consistently high in openness to experience, conscientiousness, and emotional stability across all \textit{small language models}, while they exhibited lower scores in extraversion and agreeableness. There was minimal variance observed among the \textit{small language models} in terms of their personality scores. For instance, OPT models tended to display higher emotional stability within the group, slightly lower openness trait scores, and increased conscientiousness. However, when taking a broader perspective, the metrics showed a substantial degree of similarity. Conversely, all \textit{large language models} exhibited significant variability in traits compared to \textit{small language models}. They demonstrated even greater increases in their openness to experience trait and displayed a wide range of diversity in emotional stability and agreeableness. This indicates that the latest LLMs are more influenced by prompt-induced trait activations.

\textbf{Standard Questions}: The chart in Figure \ref{fig:standard_questions_families} compares the responses of the GPT, Llama2, and Mixtral families when presented with standard questions. Each polar chart is divided into five regions, representing the five types of questions from the standard question prompt set. Within each region, five points correspond to the initial five prompts used to generate texts. It is observable that the models exhibit varying levels of sensitivity to the input prompts, with significant variations depending on the initial prompt. For GPT-3.5 and GPT-4, a discernible evolution in personality traits is noted, particularly a progressive enhancement in the degree and diversity of the Openness trait. For Standard Questions, "Plans for the future" and "Tell me about yourself" show a notable increase in this trait. In terms of Agreeableness, we observe very high scores, especially for responses related to "Strengths and Weaknesses", "handling pressure", and "Tell me about yourself" prompts. Llama2 exhibits similar personality traits, but its chatbot versions show deviations, increasing their scores in most traits except in Extraversion and demonstrating a higher sensitivity to the Agreeableness trait in responses to "Tell me about yourself". For the Mixtral family, the displayed personality traits exhibit consistency across its three versions.

In \textit{small language models}, the personality trait reflected for all standard interview questions presents a very similar behavior. Xlnet shows a slightly different because decrease their Agreeableness trait compared with the others and OPT family shows a very low value trait score for Openess and Emotional Stability. \textit{Large language models} manage to maintain their personality traits less invariant, but each model exhibits a more distinct personality from the others. Details are shown in Appendix \ref{ap:sq_small_models}. The mean Big Five score for each category is shown in Table \ref{tab:average_score}, which demonstrates a  distinct demarcation between \textit{small language models} and \textit{large language models}. The latter have increased in size to enhance their reasoning capabilities. This enhancement, coupled with an expanded dataset during the training phase and the deliberate development of models to function as assistants, aims at the construction of a tailored personality profile, thus amplifying the prevalence of beneficial traits within these systems.

\begin{table}[h]
    \centering
    \caption{Analysis Results of LLMs on Personality Traits. Scores represent the average probability of each LLM exhibiting the corresponding trait, with \textbf{the highest scores} per trait in bold and \textit{the second-highest scores} in italics.}
    \label{tab:average_score}
    \begin{tabular}{lccccc}
    \toprule
    Model & Openness & Conscientiousness & Extraversion & Agreeableness & Emotional Stability \\
    \midrule
    GPT-4-Turbo & \textit{50.79\%} & \textit{51.48\%} & 45.34\% & \textit{83.50\%} & \textit{51.30\%} \\
    GPT-3.5-Turbo & \textbf{51.91\%} & \textbf{51.84\%} & 47.58\% & \textbf{86.37\%} & \textbf{52.68\%} \\
    GPT-2-Xl & 41.95\% & 44.73\% & 42.25\% & 62.16\% & 38.04\% \\
    GPT-2-Large & 41.71\% & 44.74\% & 42.20\% & 62.00\% & 38.12\% \\
    GPT-2-Medium & 41.74\% & 44.43\% & 42.26\% & 62.67\% & 37.26\% \\
    GPT-2-Base & 42.01\% & 44.75\% & 43.00\% & 61.90\% & 38.31\% \\
    \midrule
    Llama2-70B-Chat & 48.98\% & 50.53\% & 44.99\% & 64.25\% & 46.17\% \\
    Llama2-13B-Chat & 48.29\% & 50.47\% & 45.51\% & 64.94\% & 46.30\% \\
    Llama2-7B-Chat & 48.48\% & 49.97\% & 45.72\% & 64.10\% & 44.63\% \\
    Llama2-70B & 44.94\% & 47.83\% & 46.54\% & 66.32\% & 42.01\% \\
    Llama2-13B & 44.88\% & 47.71\% & 46.55\% & 66.07\% & 42.00\% \\
    Llama2-7B & 45.24\% & 47.75\% & 46.75\% & 66.35\% & 41.87\% \\
    \midrule
    Mixtral-8X7B-Instruct & 45.61\% & 48.26\% & 46.14\% & 64.95\% & 43.96\% \\
    Mistral-7B-Instruct & 48.00\% & 49.17\% & 44.00\% & 64.54\% & 45.26\% \\
    Mistral-7B & 45.31\% & 47.50\% & 45.98\% & 64.94\% & 42.40\% \\
    \midrule
    Falcon-7B-Instruct & 42.60\% & 47.43\% & 46.72\% & 67.62\% & 42.05\% \\
    Falcon-7B & 44.05\% & 47.27\% & 46.70\% & 66.51\% & 41.79\% \\
    \midrule
    Bloomz-7B1 & 42.96\% & 47.64\% & 47.85\% & 71.29\% & 47.16\% \\
    Bloomz-3B & 43.04\% & 47.64\% & \textit{48.37\%} & 69.87\% & 46.92\% \\
    Bloomz-560M & 41.59\% & 48.48\% & \textbf{50.08\%} & 69.33\% & 47.27\% \\
    Bloom-7B1 & 43.29\% & 47.20\% & 45.85\% & 65.82\% & 41.89\% \\
    Bloom-3B & 43.27\% & 46.99\% & 46.23\% & 66.48\% & 41.99\% \\
    Bloom-560M & 43.22\% & 46.48\% & 46.32\% & 65.88\% & 41.39\% \\
    \midrule
    OPT-1.3B & 39.34\% & 42.94\% & 42.17\% & 64.12\% & 38.57\% \\
    OPT-350M & 38.18\% & 42.46\% & 42.13\% & 64.62\% & 38.45\% \\
    OPT-125M & 38.12\% & 42.38\% & 42.43\% & 63.98\% & 37.79\% \\
    \midrule
    Xlnet-Base-Cased & 46.07\% & 47.17\% & 44.80\% & 58.21\% & 40.56\% \\
    \bottomrule
    \end{tabular}
\end{table}

%\begin{table}[ht]
%    \caption{Mean Big Five score for each category}
%    \begin{center}    
    %\begin{tabular}{lccccc}\\
    %\hline
%Model & Openness & Conscientiousness & Extraversion & Agreeableness & Emotion Stability %\\ \hline \\
%XLNet-base-cased & 3.97 & 3.67 & 2.27 & 2.55 & 3.74 \\
%GPT2 & 3.84 & 3.57 & 2.39 & 2.67 & 3.77 \\
%OPT-125m & 3.95 & 3.66 & 2.51 & 2.96 & 4.03 \\
%XLNet-large-cased & 4.11 & 3.73 & 2.39 & 2.61 & 3.76 \\
%GPT2-medium & 3.94 & 3.58 & 2.40 & 2.65 & 3.64 \\
%OPT-350m & 4.10 & 3.70 & 2.55 & 3.05 & 4.11 \\
%GPT2-large & 3.97 & 3.61 & 2.43 & 2.73 & 3.71 \\
%OPT-1.3b & 4.18 & 3.72 & 2.55 & 3.00 & 4.09 \\ \hline
%Falcon-7b & 4.57 & 3.54 & 2.70 & 3.19 & 3.42 \\
%Llama-2-7b & 4.58 & 3.89 & 2.69 & 2.46 & 3.83 \\
%GPT-3.5 & 4.65 & 3.82 & 2.46 & 3.40 & 4.14 \\
%GPT-4 & 4.44 & 3.75 & 2.36 & 3.61 & 4.16 \\
%\end{tabular}\label{tab:sq_table}
    %\end{center}
%\end{table}

\subsection{Fine-tuning in \textit{small language models}}
It is discernible from our findings that the process of fine-tuning exerts a subtle influence on the model's personality traits. The specific text utilized for fine-tuning appears to play a pivotal role, with certain traits experiencing augmentation or diminution accordingly. Like with the standard models, the fine-tuned models tend to be high on Agreeableness. The results in Table \ref{tab:my_label}show that all tested models increased the agreeableness trait and decreased the conscientiousness and emotion stability (except Shakespeare) trait after the fine-tuning. Notably, the GPTJ-6B and GPT2 models exhibit similar trait scores, albeit with discernible differences Agreeableness. This variation becomes more pronounced when examining models fine-tuned with specialized datasets. For instance, the GPTJ-6B model fine-tuned with Shakespearean texts notably scores the highest in Agreeableness, possibly reflecting the linguistic style and nuances of Shakespeare's dialogues. Similarly, variants like GPT2 fine-tuned with data reflective of public figures such as Elon Musk or Michael Jackson display distinct personality characteristics, likely mirroring aspects of these individuals' public personas. These divergences in personality traits are not just academic observations; they have practical implications, particularly in how language models engage in tasks involving human interaction, be it conversational interfaces, creative text generation, or scenarios necessitating empathy and nuanced tone.

\begin{table}[ht]
    \centering
    \caption{Trait Activating Score for fine-tunning in \textit{small language models}. Scores represent the average probability of each LLM exhibiting the corresponding trait, with \textbf{the highest scores} per trait in bold and \textit{the second-highest scores} in italics.}
    \label{tab:my_label}
    \begin{tabular}{lccccc}
    \toprule
    Model & Openness & Conscientiousness & Extraversion & Agreeableness & Emotional Stability \\
    \midrule
    GPT2 & \textit{48.22\%} & \textit{43.96\%} & 42.15\% & 62.62\% & 33.95\% \\
    \midrule
    GPT2/Shakespeare & 58.30\% & \textbf{44.81\%} & 43.87\% & \textit{79.03\%} & \textit{40.56\%} \\
    GPT2/Elon Musk & 44.03\% & 43.83\% & 43.84\% & 74.22\% & 39.52\% \\
    GPT2/Michael Jackson & 55.77\% & 40.73\% & 44.03\% & 89.18\% & 6.77\% \\
    GPT2/Rihanna & 45.03\% & 41.84\% & \textbf{50.24\%} & 80.88\% & 23.21\% \\
    GPT2/Yan Lecun & 52.02\% & 41.52\% & 36.17\% & 70.82\% & 32.84\% \\
    \midrule
    GPTJ-6B & 47.35\% & 43.31\% & 42.37\% & 56.89\% & 33.57\% \\
    \midrule
    GPTJ-6B/Shakespeare & \textbf{63.22\%} & 41.53\% & 43.89\% & \textbf{100.00+\%} & \textbf{42.11\%} \\
    GPTJ-6B/4-Chan & 44.01\% & 35.94\% & \textit{48.35\%} & 58.99\% & 23.62\% \\
    GPTJ-6B/Shinen & 37.78\% & 37.78\% & 40.32\% & 59.99\% & 30.84\% \\
    \bottomrule
    \end{tabular}
\end{table}

\section {Discussion}
This exploratory study investigated the personality of 27 autoregressive language models. Through providing the language models with question prompts to complete (due to limitations with question and answering capabilities) that reflect questions asked in job interviews, the Big Five personality profile of the neural networks was analysed based on the output text. By analysing the text using the Personality Prediction from our fine-tuned classifiers, initially created to predict personality from Facebook statuses, we found all \textit{small language models} are high in openness to experience, even when trait-activating questions were used to attempt to elicit higher levels of other traits. Conscientiousness, agreeableness and emotional stability were relatively similar across the models and were present at medium levels, scores for extraversion were consistently lower.There are multiple reasons that could have resulted in these high levels of openness to experience. For example, the models were trained on Wikipedia, books, news articles etc., all of which are intellectual and informative materials. Therefore, this training data may have been high in openness, resulting in the  models also reflecting this. Another reason could be that the model for predicting personality was most accurate for openness. Indeed, this consistent with prior research that indicates that openness is the easiest trait to infer from text \cite{golbeck2011predicting} \cite{poria2013common}. To further investigate the source of high levels of openness to experience, future research could examine personality at a facet level; we propose that high levels of openness are due to high levels of intellect, but future research could investigate this. Although \textit{large language models} exhibit a broader range of personality traits, they remain impervious to trait-activation, with the most significant variance in mean trait scores across two standard questions being a mere 0.4 (on a 1 to 5 scale). This minimal fluctuation persists between standard interview and trait-activating questions, contradicting previous studies involving human subjects which showed amplified traits when trait-activation methods were employed \cite{tett2003personality} \cite{lievens2006large} \cite{speer2015assessment} \cite{holtrop2022exploring}\cite{hickman2021automated}. This discrepancy underscores a fundamental divergence between human responses and neural network outputs, the latter being unaffected by social nuances. This phenomenon may be attributed to the fact that the lower log probability observed in human compositions, indicative of a nuanced and varied manifestation of personality, is something that high-probability-focused language models fail to replicate, resulting in text outputs with more homogenized and limited personality reflections. Such limitations fuel apprehensions regarding AI's role in recruitment, as there's a perceived absence of human touch when algorithms assess candidates' performances. To this end, future research could examine the effectiveness of the trait-activating questions by using the same approach with human participants, where they would be asked to complete the question prompts for both the trait-activating questions. Within group differences could then be examined to investigate whether levels of each trait are increased through trait-activation. Between group differences could also be examined to compare the scores for the human participants with the neural networks. The findings of this study have implications for the use of generative AI by job applicants in the interview process. Indeed, reliance on conversational AI to prepare answers to interview question could influence the way that they are perceived, and also result in misalignment with applicants' presentation and their true personality traits. This, in turn, could impact the utility of the interviews if applicants' true profiles cannot be identified since predictions about job performance could be inaccurate. However, given that the models can lack variability in their outputs, this could provide an avenue to identify applicants' use of generative AI by comparing the similarity of answers to questions, particularly for recorded interviews \cite{hirevue2023}.
573

\section{Conclusion}
This study sought to investigate the personality profiles of language models using a text-based classifier approach using prompts derived from recruitment interviews. Specifically, 25 of the 50 prompts were designed to reflect common interview questions, and the remaining 25 were designed to elicit particular personality traits using trait-activation (5 per trait).  In general, the language models had high levels of openness, low extraversion, and moderate levels of the other three Big Five traits (agreeableness, emotional stability, and conscientiousness). Unlike humans, these models are not influenced by trait-activation, likely due to the absence of social cues in computational models. Models such as Falcon, Llama, GPT (3.5 and 4) display a broader spectrum of traits but remain unaffected by trait-activation, further enhancing traits like openness, agreeableness, and emotional stability. This study's approach of using dual prompts to elicit trait-specific responses sheds light on AI's functioning, enhancing its transparency and explainability. Such insights are valuable in recruitment contexts to discern and regulate the use of AI, maintaining the integrity of hiring decisions.

\section{Ethical considerations}
This study did not involve human participants and therefore raises no ethical issues in terms of physical and psychological harm. 

\section{Researcher positionality}
This study was conducted by an interdisciplinary team of researchers who have both applied industry knowledge and strong academic foundations and affiliations. Namely, the research team is formed of three computer scientists and a I-O psychologist who all research ethical AI. Moreover, multiple members of the team specialise in NLP and LLMs and have worked in the domain of algorithmic and AI-driven recruitment tools.

Our novel approach to personality trait elicitation is a result of our combined experiences, where direct experience in the recruitment field led to the development of the prompts designed to be reflective of interview questions. Moreover, expertise in text generation and text analytics meant that the outputs of the models could be used to investigate the personality of the models, as opposed to the approach of other researchers involving providing the models with scales to respond to scoring their responses.

\section{Adverse impact statements}
While this experimental study sought to provide greater explainability around LLMs and how they interact, it is important to consider the implications the findings of this paper could have on the hiring process. For example, applicants high in openness to experience, for example, may be subject to accusations of "cheating" using generative AI even if they did not use it. Nevertheless, this study aims to have a positive impact on the fields of computer science, I-O psychology, and human resource management by providing an insight into how the use of generative AI tools to prepare for interviews may affect applicant personality inferences and profiles and allowing more informed decisions to be made about their use.

\begin{acks}
\end{acks}

\bibliographystyle{ACM-Reference-Format}
\bibliography{reference}

\appendix

\section{Question stems}\label{ap:questions}
\subsection{Standard Interview Questions}

Tell me about yourself:
\begin{itemize}
\item I would describe myself as
\item Others would describe me as
\item My hobbies are
\item After work, I like to
\item At the weekend, I usually
\end{itemize}

Cultural fit:
\begin{itemize}
\item My idea workplace would be 
\item I think that working as a team is
\item When I am involved in a conflict in the workplace, I  
\item I think that workplace socials are 
\item When I am in the office, I like to
\end{itemize}

Strengths and weaknesses/challenges: 
\begin{itemize}
\item My strengths are
\item My weaknesses are 
\item When presented with a challenge, I 
\item I am skilled at 
\item My best quality is
\end{itemize}

Plans for the future (where do you see yourself in X years):
\begin{itemize}
\item When I think of the future, I 
\item I think that career progression is
\item In the future, I would like to 
\item I think that the future will be
\item Through my work, I hope to 
\end{itemize}

Coping under pressure: 
\begin{itemize}
\item When I have a deadline coming up, I 
\item If I am working on many projects at once, I
\item Being able to priorities tasks is 
\item When I feel stressed at work, I 
\item When working under pressure, I typically
\end{itemize}

\subsection{Trait-activating questions: (C, E and ES adapted from \cite{holtrop2022exploring})}
Openness:
\begin{itemize}
\item Learning new information makes me
\item I think that art is 
\item When I go on vacation, I typically
\item Trying new foods is something that I
\item I think that daydreaming is
\end{itemize}

Conscientiousness:
\begin{itemize}
\item I think that perfectionism is
\item To motivate myself to work, I usually
\item When it comes to being organised, I typically
\item I think that having work-related goals is 
\item When making decisions, I typically
\end{itemize}

Extraversion: 
\begin{itemize}
\item I think that being active is 
\item During a social situation, I think of myself as
\item When I am in charge of others, I feel 
\item When I am with a group of people, I 
\item When I am alone, I
\end{itemize}

Agreeableness:
\begin{itemize}
\item When I achieve something, others should 
\item When someone needs help, I
\item I think that rules are
\item Confrontations with others are
\item I feel sympathy for
\end{itemize}

Emotional stability:
\begin{itemize}
\item When I encounter a stressful situation, I 
\item Being the center of attention makes me feel
\item My mood most of the time is 
\item My opinion of myself is
\item When I am craving something, I usually
\end{itemize}
\newpage

\section{Standard and Trait Activating Questions}

\begin{figure}[!htbp]
    \centering

    \begin{subfigure}{\linewidth}
        \centering
        \includegraphics[width=\linewidth, keepaspectratio]{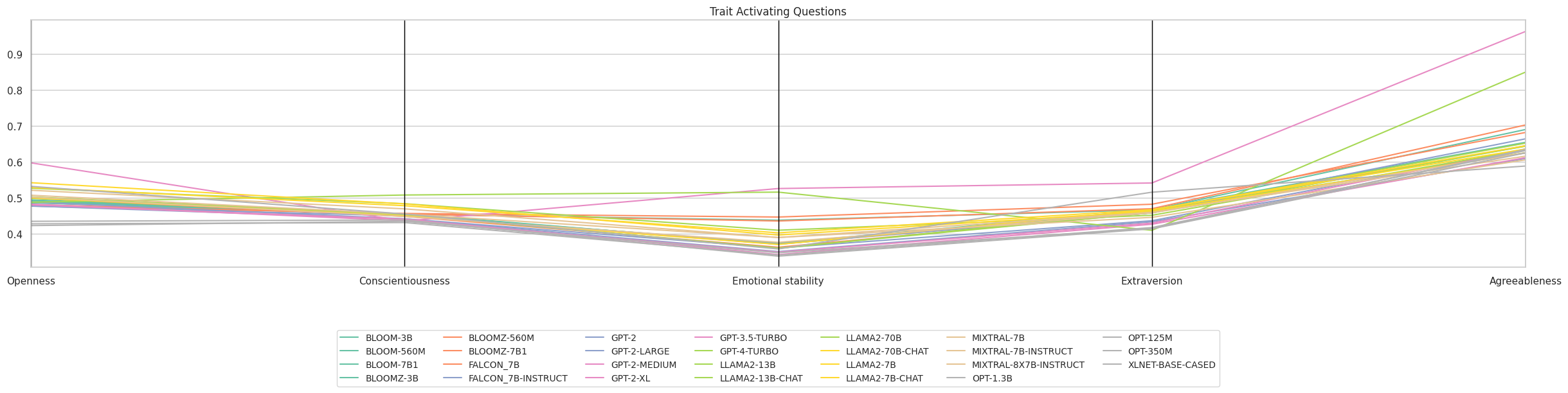}
        \caption{Personality Trait Score for Trait Activating Questions.}
        \label{fig:ptrait_act}
    \end{subfigure}

    \begin{subfigure}{\linewidth}
        \centering
        \includegraphics[width=\linewidth, keepaspectratio]{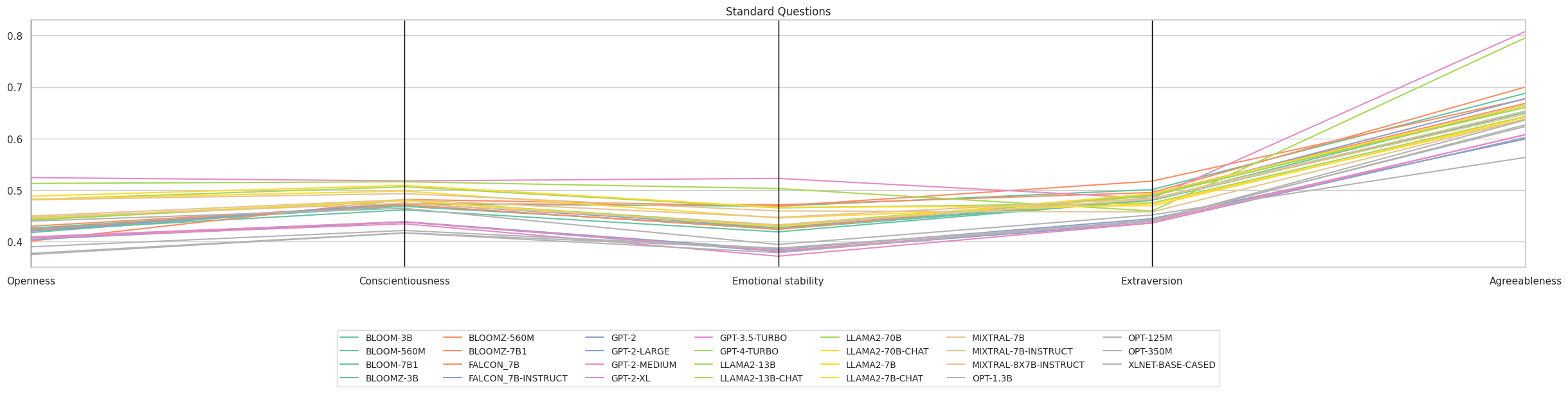}
        \caption{Personality Trait Score for Standard Questions.}
        \label{fig:ptrait_std}    
    \end{subfigure}
    
    \caption{Personality Trait score for Trait Activating and Standard Questions}
    \label{fig:personality_trait_act_std}
\end{figure}

\section{Standard Questions in \textit{Small language Models}}\label{ap:sq_small_models}

In a comprehensive analysis, observing Figure \ref{fig:aq_small_models} it becomes evident that \textit{smaller models} exhibit a remarkably consistent response pattern, regardless of the nature or type of questions presented. Our data reveals a pronounced surge in both emotional stability and agreeableness traits across the three distinct variants of the OPT model, namely OPT-125m, OPT-350m, and OPT-1.3B. 

This increase is especially evident when these models are presented with questions such as "Tell me about yourself," delve into "Culture Fit," or explore an individual's "Plans for the future." However, all models show notable variability in the 'openness' trait, which seems tied to the question's theme. For instance, with "Plans for the future," they emphasize 'openness,' yielding a median rating near 5. But with "Tell me about yourself," the openness score drops to an average median of 3. Regarding "Plans for the future," these models highlight traits like Openness, Agreeableness, and Extraversion.

\begin{figure}[H]
  \centering
  \begin{tabular}{cc}
  \includegraphics[width=\linewidth]{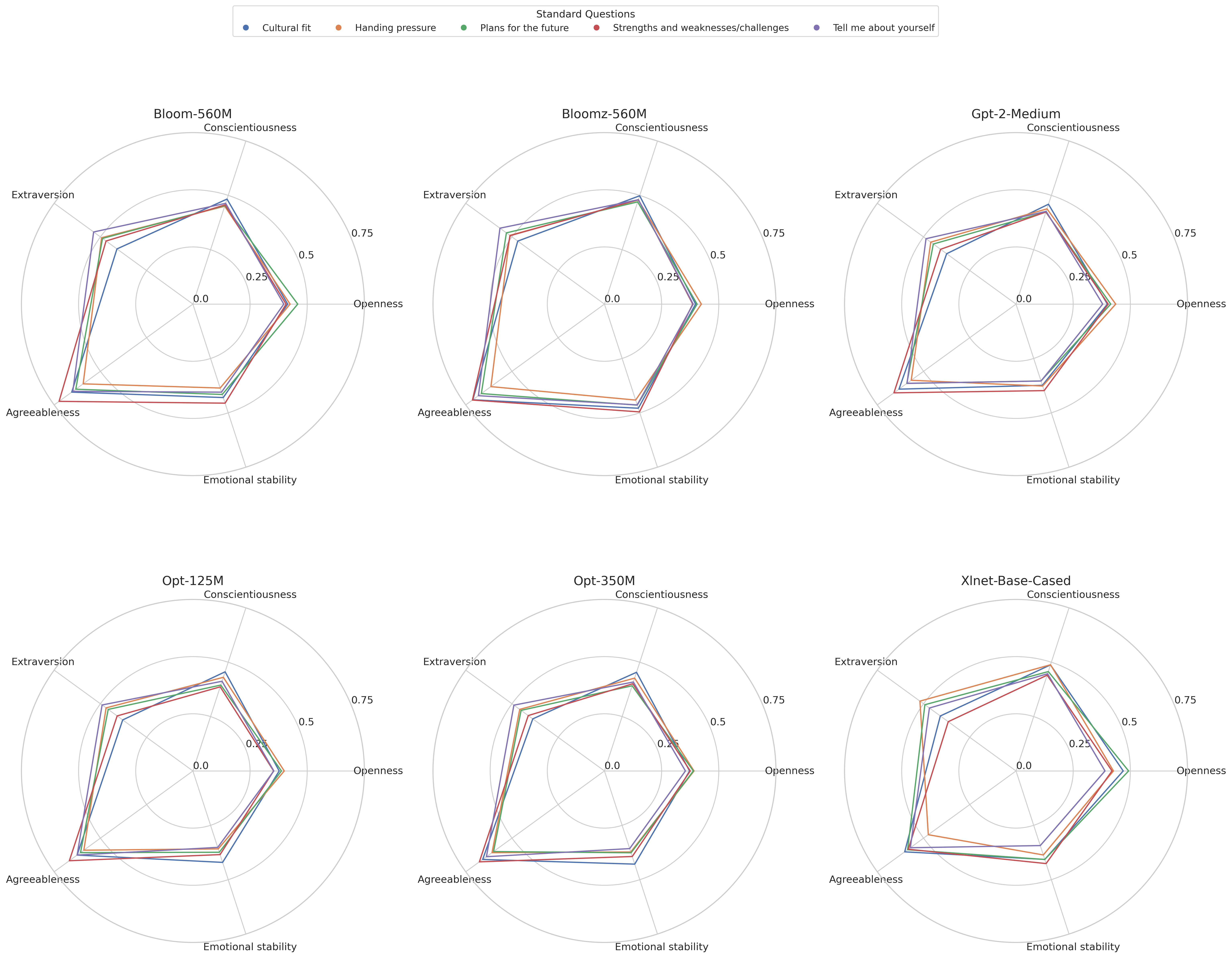}
  \end{tabular}
  \caption{Comparison of the Mean Big Five Scores for each for categories of standard interview questions.}
  \label{fig:aq_small_models}
\end{figure}

\begin{figure}[H]
  \centering
  \begin{tabular}{cc}
    \includegraphics[width=\linewidth]{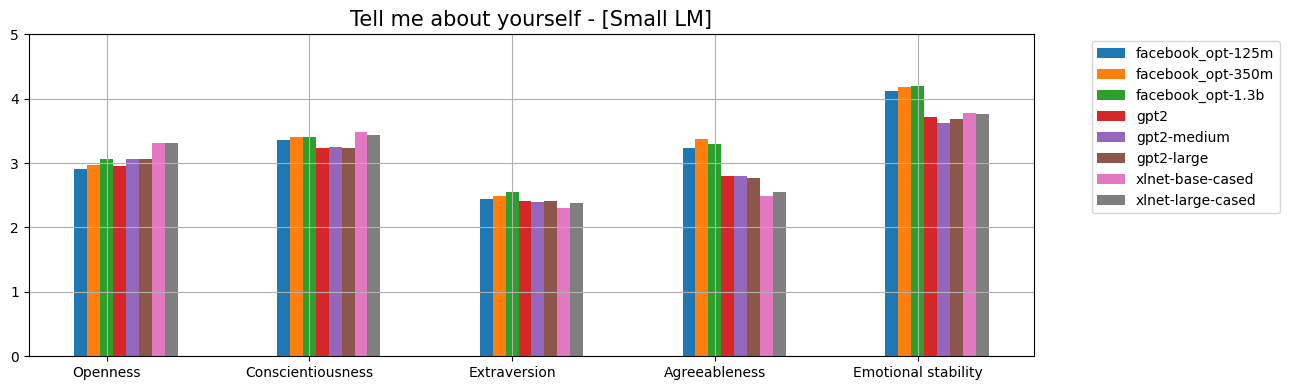} \\ \includegraphics[width=\linewidth]{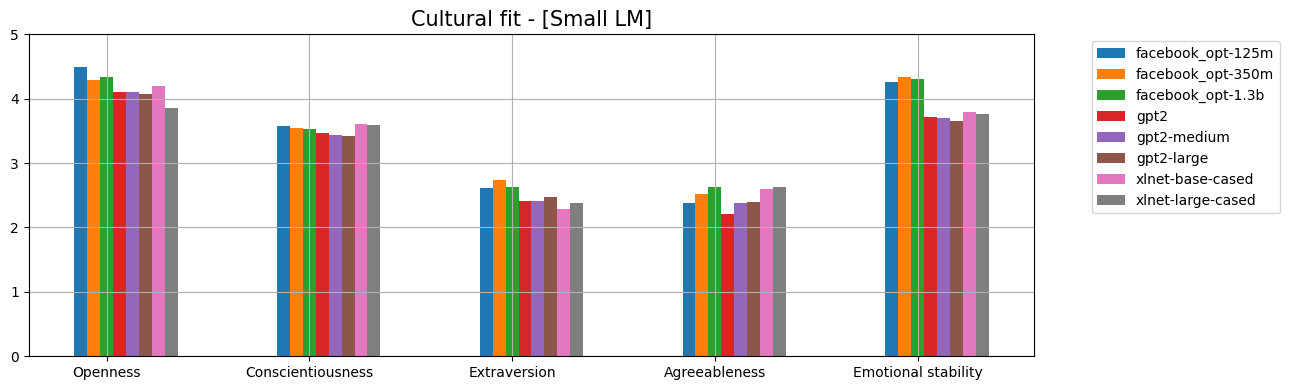} \\
    \includegraphics[width=\linewidth]{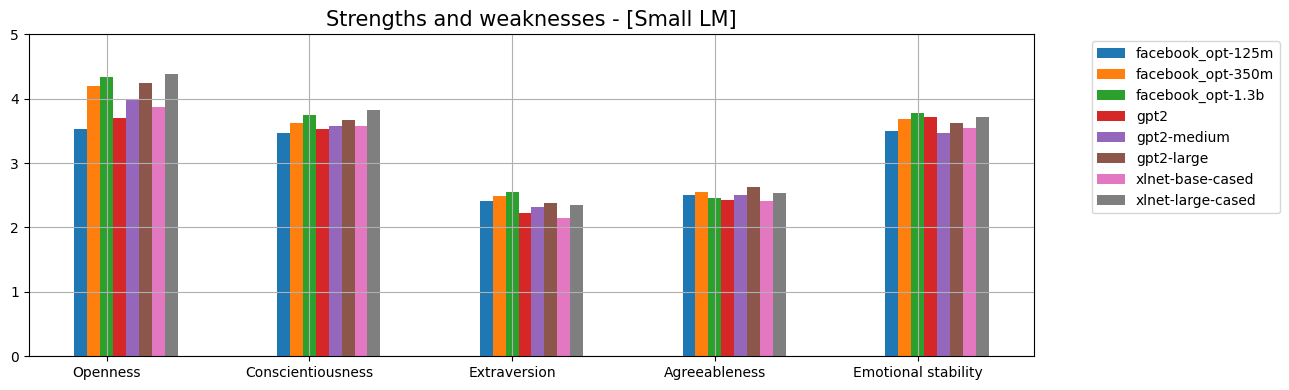} \\ \includegraphics[width=\linewidth]{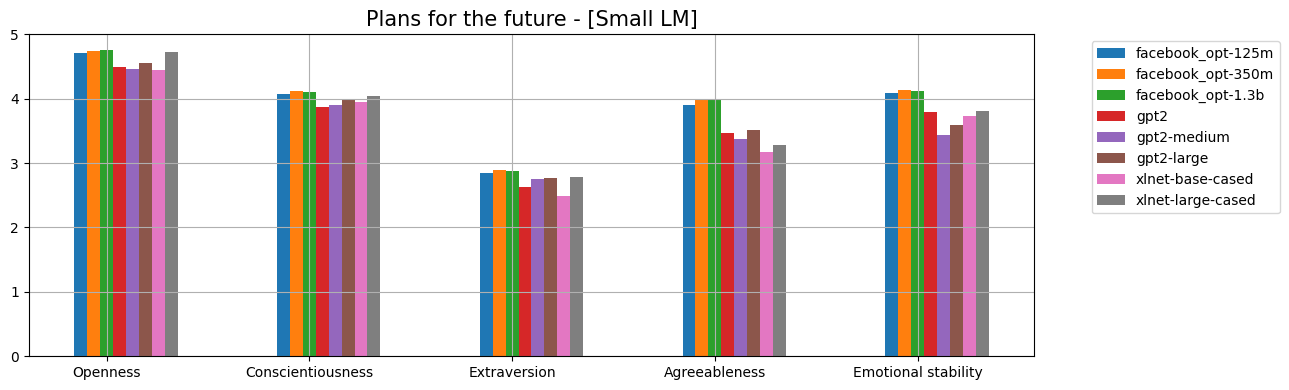}
  \end{tabular}
  \caption{Comparison of the Mean Big Five Scores for each for categories of standard interview questions.}
  \label{fig:aq_small_models}
\end{figure}

\begin{figure}[H]
  \centering
    \includegraphics[width=\linewidth]{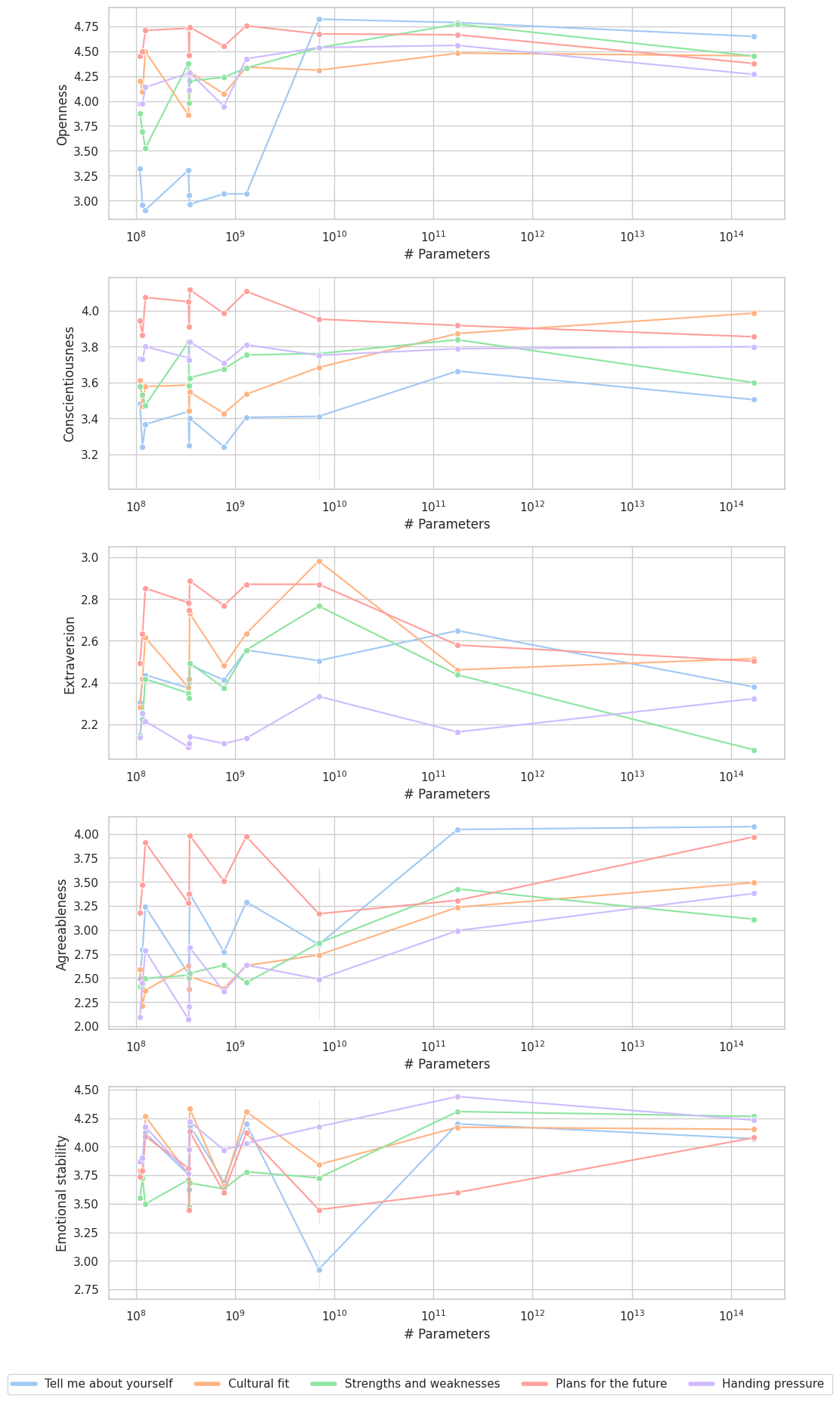}
  \caption{Comparison of the Mean Big Five Scores to the Number of Parameters. Each plot illustrates the response for individual categories of standard interview questions}
  \label{fig:sq_trait_vs_num_parameters}
\end{figure}

In Figure \ref{fig:sq_trait_vs_num_parameters}, we can distinctly see the relationship between personality traits in language models and their number of parameters. Larger models, which also happen to be the more recent iterations, have consistently shown increased scores across all questions in traits such as Openness, Agreeableness, and Emotional Stability. Furthermore, when assessing questions related to 'Strengths and Weaknesses', the GPT-4 model exhibits an Extraversion score around 2, a considerably low value compared to any of its predecessors. Similarly, another notable observation is that GPT-3.5 displays a more pronounced trait of emotional stability than GPT-4 for four out of the five questions. Interestingly, "Plans for the future" is the only question where GPT-3.5 scores lower than GPT-4."

\section{Performance of Composite Models}\label{ap:fine-tunning}

\begin{figure*} [!h]
    \centering
    \includegraphics[width=\linewidth]{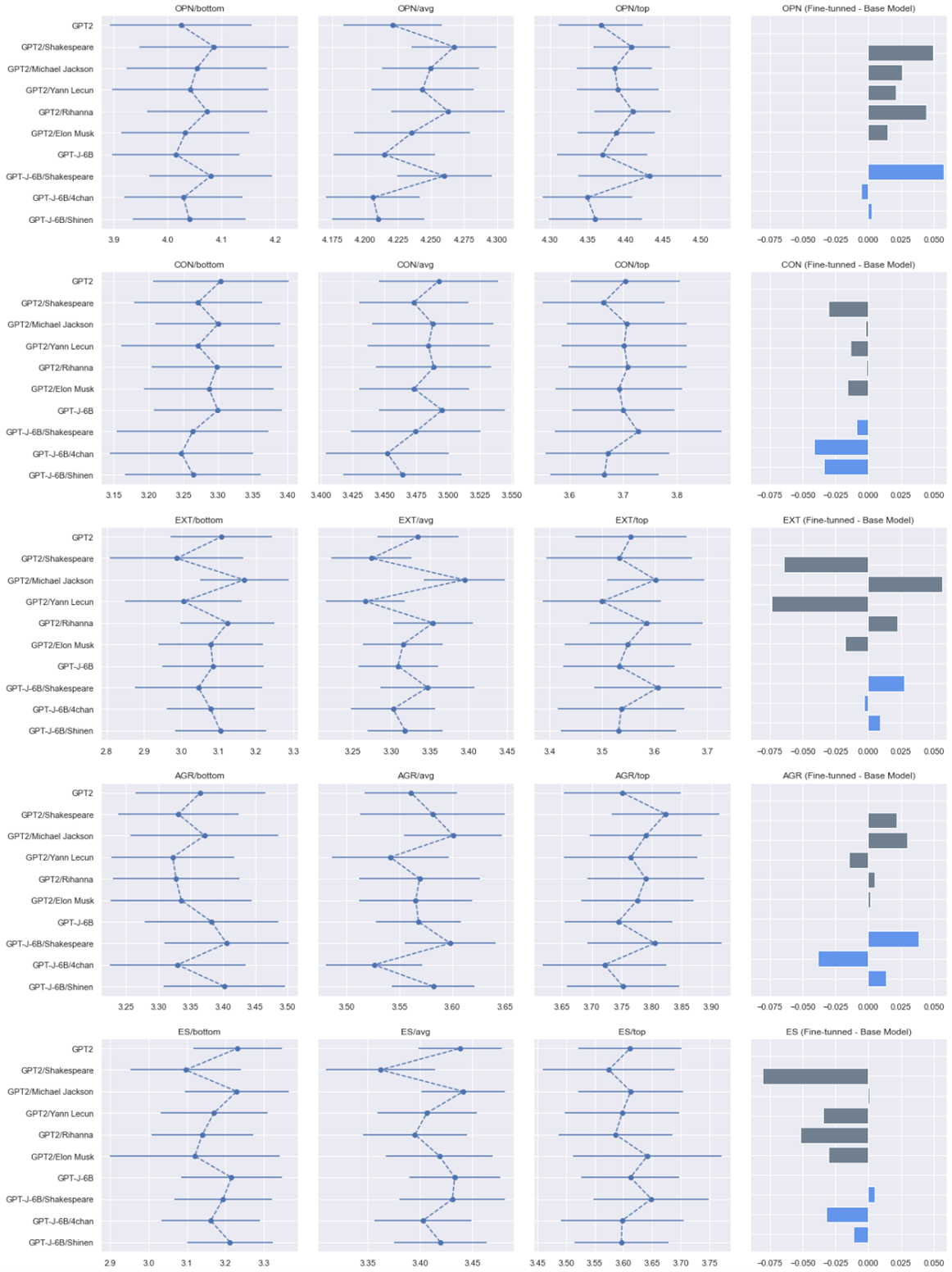}
    \caption{The figure displays the changes in personality traits in the model after undergoing fine-tuning. The gray colors in the 4th column represent the changes for the base GPT-2 model, while the light blue colors indicate the changes using the GPT-J base model.}
    \label{fig:my_label}
\end{figure*}

\begin{figure}
    \centering
    \begin{subfigure}{\linewidth}
        \centering
        \includegraphics[width=0.95\linewidth]{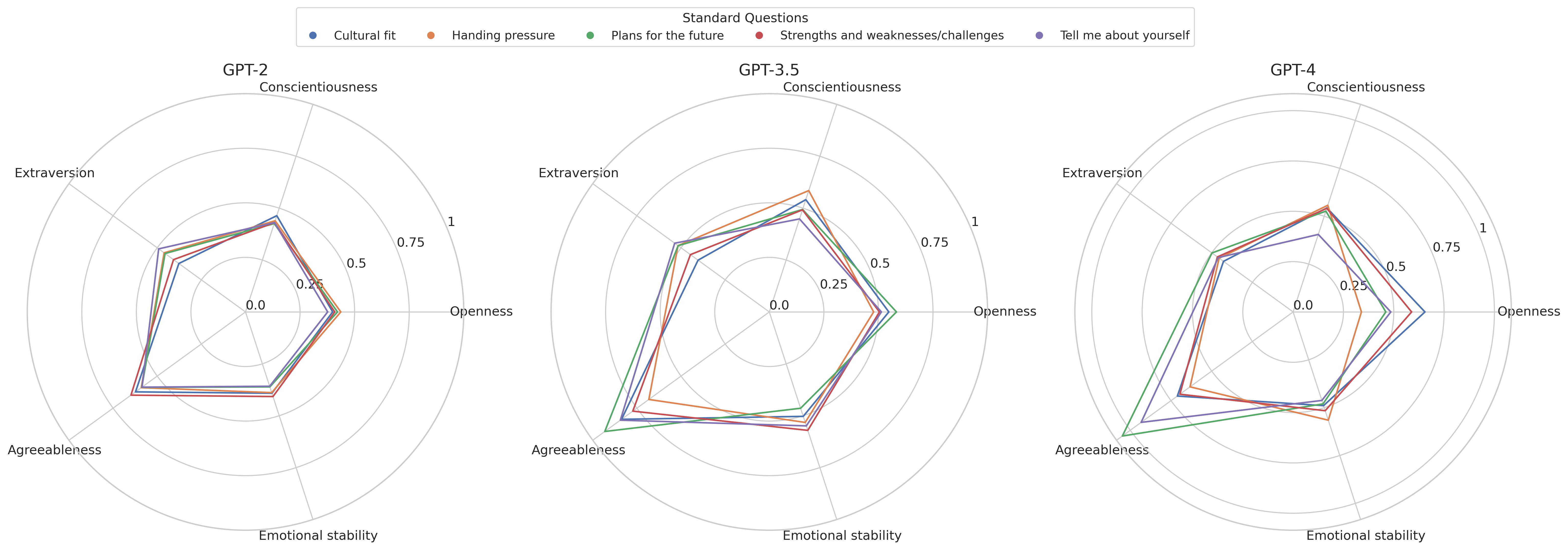}
        \caption{Personality Trait Score for GPT Family.}
        \label{fig:standard_questions_gpt}
    \end{subfigure}
    
    \begin{subfigure}{\linewidth}
        \centering
        \includegraphics[width=0.95\linewidth]{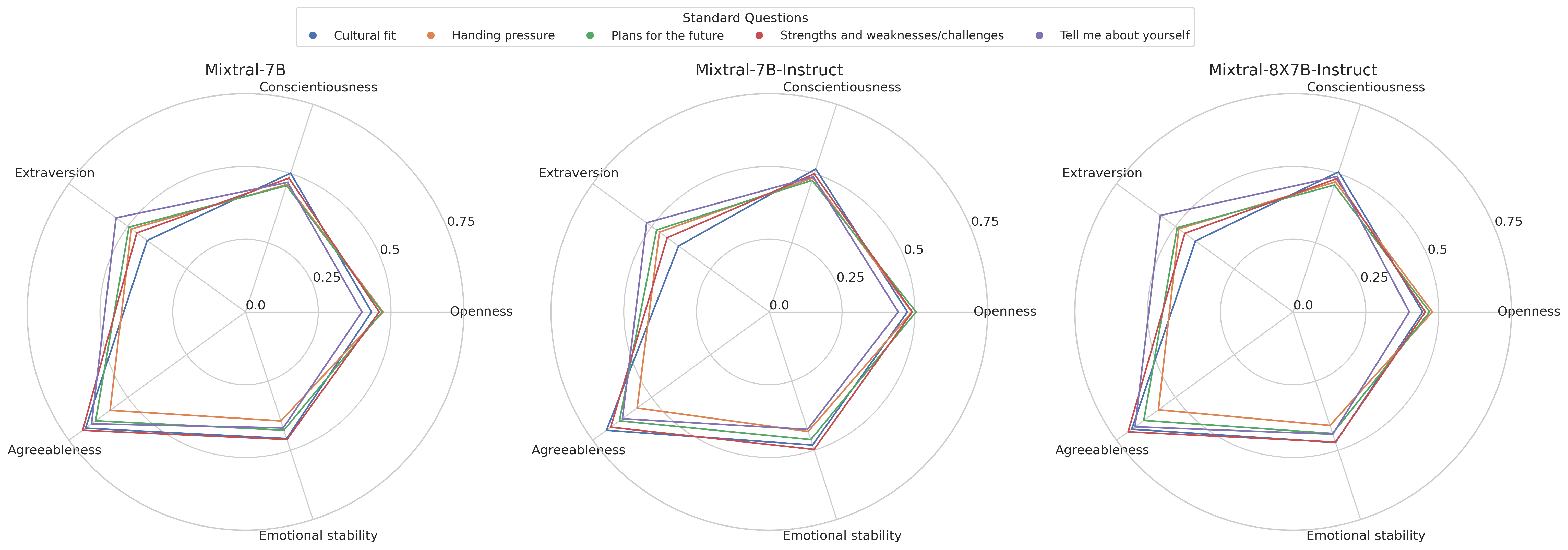}
        \caption{Personality Trait Score for Mistral Family.}
        \label{fig:standard_questions_mistral}
    \end{subfigure}

    \begin{subfigure}{\linewidth}
        \centering
        \includegraphics[width=0.95\linewidth]{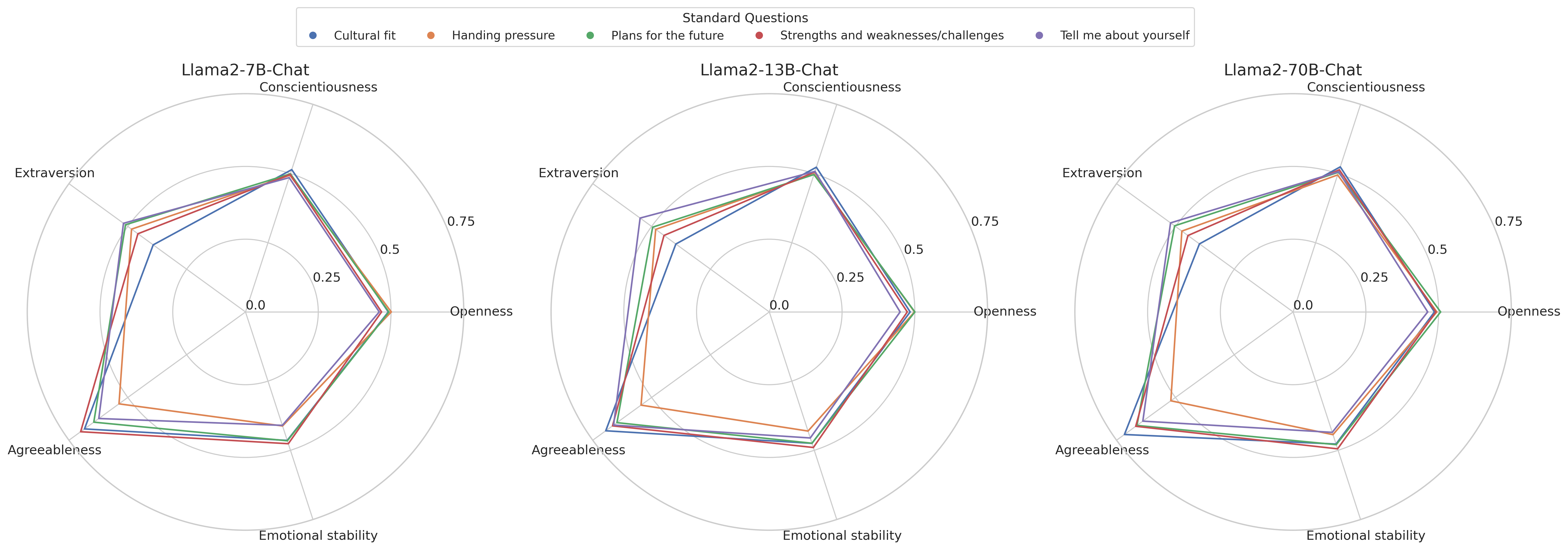}
        \caption{Personality Trait Score for Llama2 Chat Family.}
        \label{fig:standard_questions_llama2chat}
    \end{subfigure}

    \begin{subfigure}{\linewidth}
        \centering
        \includegraphics[width=0.95\linewidth]{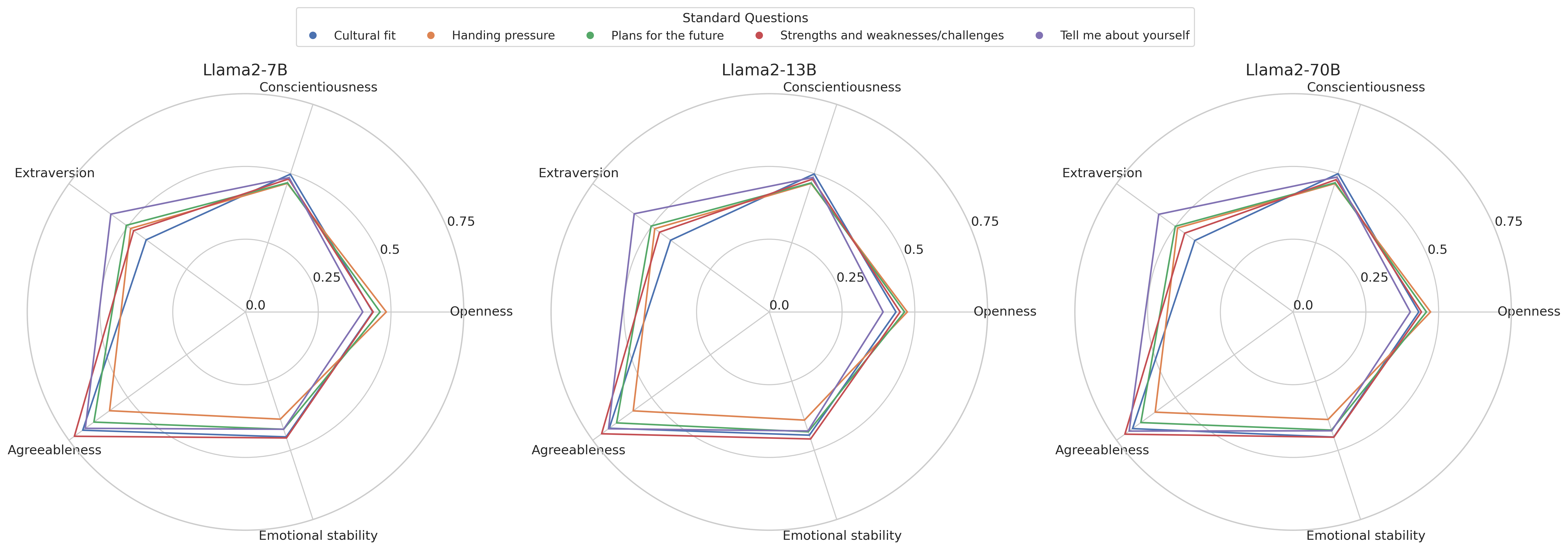}
        \caption{Personality Trait Score for Llama2 Base Family.}
        \label{fig:standard_questions_llama2}
    \end{subfigure}
    
    \caption{Comparison of the Mean Big Five Scores for each for categories of standard interview questions.}
    \label{fig:combined}
\end{figure}









\end{document}